%% file: main.tex
\documentclass[conference]{IEEEtran}
\IEEEoverridecommandlockouts
\usepackage{cite}
\usepackage{amsmath,amssymb,amsfonts}
\usepackage{algorithmic}
\usepackage{graphicx}
\usepackage{textcomp}
\usepackage{xcolor}
\usepackage{float}
\usepackage{rotating}
\usepackage{upgreek}
\usepackage{mathtools}
\usepackage{diagbox}

\usepackage{graphicx}
\usepackage[font={sf,scriptsize},           
            labelfont={bf,color=accessblue},
            caption=false]                  
           {subfig}

\usepackage[ruled,vlined,linesnumbered]{algorithm2e}
\SetKwProg{Init}{init}{}{}

\usepackage{amsmath}

\usepackage{comment}

\newsavebox\mybox

\def\BibTeX{{\rm B\kern-.05em{\sc i\kern-.025em b}\kern-.08em
    T\kern-.1667em\lower.7ex\hbox{E}\kern-.125emX}}
\begin{document}

\title{


DEK-Forecaster: A Novel Deep Learning Model Integrated with EMD-KNN for Traffic Prediction


}

\author{\IEEEauthorblockN{Sajal Saha, Sudipto Baral, and Anwar Haque}
\IEEEauthorblockA{\textit{Department of Computer Science} \\
\textit{University of Western Ontario, London, ON, Canada}\\
Email:\{ssaha59, sbaral3, ahaque32\}@uwo.ca}
}

\maketitle

\begin{abstract}
Internet traffic volume estimation has a significant impact on the business policies of the ISP (Internet Service Provider) industry and business successions. Forecasting the internet traffic demand helps to shed light on the future traffic trend, which is often helpful for ISPs’ decision-making in network planning activities and investments. Besides, the capability to understand future trend contributes to managing regular and long-term operations. This study aims to predict the network traffic volume demand using deep sequence methods that incorporate Empirical Mode Decomposition (EMD) based noise reduction, Empirical rule based outlier detection, and $K$-Nearest Neighbour (KNN) based outlier mitigation. In contrast to the former studies, the proposed model does not rely on a particular EMD decomposed component called Intrinsic Mode Function (IMF) for signal denoising. In our proposed traffic prediction model, we used an average of all IMFs components for signal denoising. Moreover, the abnormal data points are replaced by $K$ nearest data point’s average, and the value for $K$ has been optimized based on the KNN regressor prediction error measured in Root Mean Squared Error (RMSE). Finally, we selected the best time-lagged feature subset for our prediction model based on AutoRegressive Integrated Moving Average (ARIMA) and Akaike Information Criterion (AIC) value. Our experiments are conducted on real-world internet traffic datasets from industry, and the proposed method is compared with various traditional deep sequence baseline models. Our results show that the proposed EMD-KNN integrated prediction models outperform comparative models.
\end{abstract}


\input{introduction.tex}

\input{literature.tex}

\input{methodology}

\input{result}

\section{Conclusion}
\label{Conclusion}
Traffic volume forecasting is an essential tool for the ISP industry to assist them in their network capacity planning activities and network investment decisions. Assessing the network traffic trend accurately helps ISPs to define, develop, and adjust their current and new infrastructure and services. Therefore, it is worthwhile to improve the accuracy of internet traffic volume predictions. This study proposes a deep learning methodology that integrates an EMD-based noise reduction and an empirical rule-based outlier detection module. Most of the previous hybrid models use EMD to obtain ensemble prediction models. Unlike the earlier studies, the proposed algorithm is not an ensemble model and does not depend on a specific Intrinsic Mode Function (IMF) for model learning. The proposed algorithm applies EMD method for denoising the original signal to grasp the general tendency of the data. After EMD denoising, the deep learning model is trained on the noise-free dataset. However, the EMD process requires the selection of a stopping criterion to determine the number of IMFs to be extracted. The choice of this criterion can significantly affect the quality of the decomposition and the effectiveness of noise reduction. In future, we plan to explore other methods of noise reduction such as Singular value decomposition (SVD)-based methods, Non-local means (NLM)-based methods, and deep learning-based methods. We identified the point outlier based on the empirical rule, and these points are mitigated with near $K$ values, optimized based on KNN regressor. There are few limitations of using of KNN for parameter optimization. For example, the KNN algorithm can be computationally expensive when the dataset has a large number of features or dimensions. As the number of features increases, the distance between the nearest neighbors can become more similar, which can make it difficult to identify the $K$ nearest neighbors. Also, the choice of distance metric can have a significant impact on the quality of the imputed values, and different distance metrics may be more appropriate for different types of data. Results are evaluated with widely used MAPE and mean accuracy measures to perform a favorable comparison. The proposed method is also compared with traditional statistical and deep sequence models and is trained on the original signal. According to the results, the proposed method outperforms all baseline prediction models. The performance of our proposed algorithm clearly shows its potential in accurately forecasting internet traffic demand compared to the other approaches.

\bibliographystyle{IEEEtran}
\bibliography{main}

\vspace{12pt}

\end{document}

%% file: introduction.tex
\section{Introduction}
\label{introduction}
With the growth and widespread use of the Internet and associated digital services and applications, there are an increasing number of different Internet business models, network traffic volumes, and network management challenges. These events present new difficulties in guaranteeing network service quality, enhancing network load capacity, fully utilizing network resources, and enhancing user experience \cite{intro1}. Network traffic is perhaps the most basic statistic used to estimate the operational health of each access point and serves as a fundamental performance indicator for the system. Numerous new difficulties, including poor network throughput and challenging network monitoring, have been brought on by the network's complexities \cite{intro2}. An essential component of network management is the amount of network activity that network equipment experiences over time. The network management strategy can be proactively altered to maximize network resources if the traffic pattern of network components can be forecasted \cite{intro4}. Moreover, traffic forecasting can be helpful in improving overall customer experience and networking quality of service (QoS) significantly.

\begin{figure}[h]
    \centering
    \includegraphics[height=4cm, width=9cm]{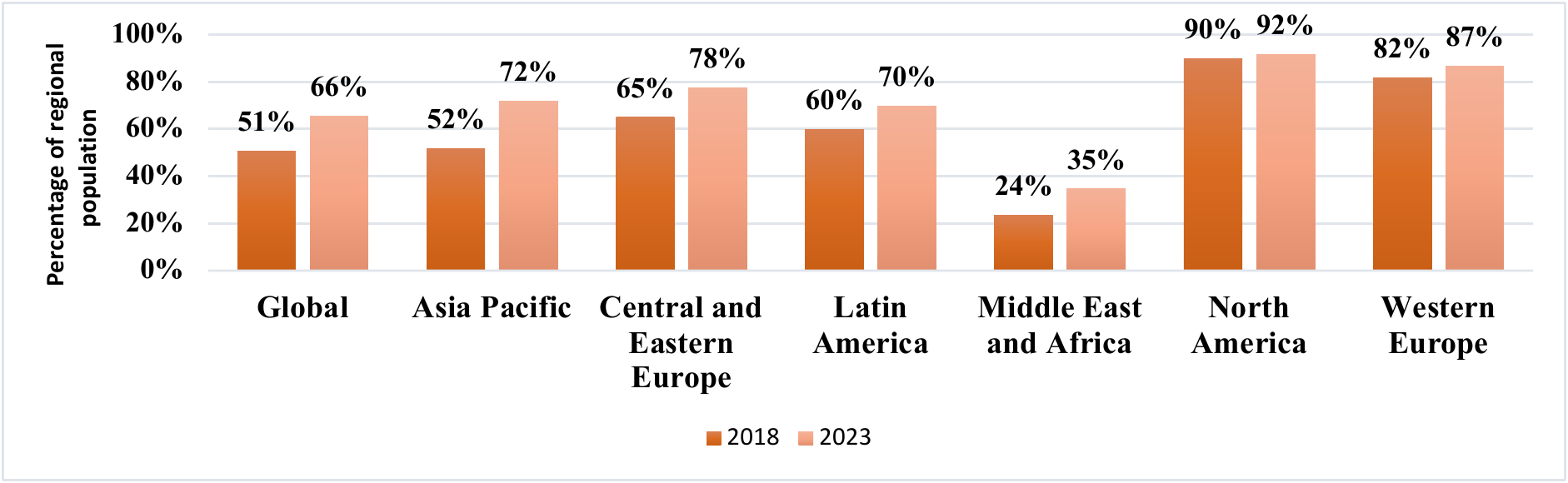}
    \caption{Regional increase in internet traffic user. }
    \label{fig:user_growth}
\end{figure}

The increasing reliance on digital communication and online services has led to a massive growth in Internet traffic, which has created significant challenges for Internet Service Providers (ISPs) in managing their networks. According to projections in the study of Cisco \cite{ciscoCiscoAnnual}, the total number of internet users worldwide is expected to increase from 3.9 billion in 2018 to 5.3 billion by 2023, which corresponds to a compound annual growth rate (CAGR) of 6\%. This means that in 2018, about 51\% of the world's population was using the internet, while by 2023, it is estimated that about 66 percent of the global population will have access to the internet. Although the growth of internet users is a worldwide phenomenon, there are regional disparities, as indicated in Fig \ref{fig:user_growth}.

While North America (and subsequently Western Europe) is projected to have the highest adoption rate throughout the forecast period, the Middle East and Africa are expected to experience the fastest growth, with a projected CAGR of 10\% from 2018 to 2023. Therefore, accurate forecasting of internet traffic volume is essential for ISPs to make informed decisions in network planning activities and investments. However, existing forecasting models often suffer from limitations, such as insufficient accuracy or the inability to handle the complexity and variability of internet traffic patterns. Previous research has focused on various techniques for predicting internet traffic volumes, such as statistical models \cite{alghamdi2019forecasting}, machine learning algorithms \cite{tang2019traffic}, and deep learning models \cite{jaffry2020cellular}. Despite these efforts, there is still a need for more accurate and reliable forecasting methods that can adapt to the dynamic nature of internet traffic. 

This study aims to address this gap by developing a novel forecasting model that combines Empirical Mode Decomposition (EMD) and K-Nearest Neighbors (KNN) methods with deep learning techniques to achieve higher accuracy and robustness in predicting internet traffic volume. This study is significant because it addresses a critical problem faced by ISPs, who need to forecast internet traffic accurately. By improving the accuracy of internet traffic forecasting, ISPs can optimize their network performance, reduce operational costs, and better meet customer demands. Additionally, this study contributes to the field of deep learning and signal processing by proposing a novel approach that integrates EMD-KNN methods with deep sequence models. Our results show that the proposed model outperforms traditional statistical and deep sequence baseline models. The potential implications of this research are far-reaching, as it can inform the development of more accurate and reliable internet traffic forecasting models for ISPs. The proposed approach can be applied to other related fields, such as predicting energy demand, traffic flow, and financial market trends. Overall, this study advances our understanding of the research problem and provides a valuable contribution to the field of deep learning and signal processing. 

As a unique time-series, network traffic reflects the interaction and influence between network services through complex features like nonlinearity, fractality, bursts, disorder, and heterogeneity \cite{intro6}. In this work, we proposed a traffic prediction methodology integrating empirical mode decomposition (EMD) based denoising and Empirical rule-based outlier detection. There are several works where EMD-based hybrid models have been proposed for traffic prediction. Most of them used EMD to decompose a signal into several components, where each component was modeled separately using either a linear or non-linear model. This multiple-model prediction strategy is time-consuming, and selecting a suitable model for a particular component is non-trivial. In this study, we aim to predict network traffic volume demand using single deep sequence methods that incorporate Empirical Mode Decomposition (EMD) based noise reduction, Empirical rule based outlier detection, and K-Nearest Neighbour (KNN) based outlier mitigation. We used real-world internet traffic datasets from industry for our experiments and compared the proposed EMD-KNN integrated prediction models with various statistical and traditional deep sequence baseline models.
Our methodology involved several steps. First, we performed EMD-based noise reduction on the dataset to remove any high-frequency noise that might affect the accuracy of our predictions. We then applied Empirical rule based outlier detection to identify any abnormal data points in the dataset. These data points were then replaced by the $K$ nearest data point's average using KNN-based outlier mitigation. To identify the best value for $K$, we performed a grid search algorithm based on the KNN regressor. We also used AutoRegressive Integrated Moving Average (ARIMA) and Akaike Information Criterion (AIC) values to select the best time-lagged feature subset for our prediction model. Finally, we used deep learning techniques to develop our EMD-KNN Traffic Forecaster model, which combines EMD-KNN methods with a deep neural network to predict network traffic volume demand accurately. The main contribution of this work is as follows:

\begin{enumerate}

    \item We introduce a novel method that combines Empirical Mode Decomposition (EMD) and K-Nearest Neighbors (KNN) with deep learning techniques to create a more accurate and robust internet traffic volume forecasting model. This model significantly improves upon the predictive performance of traditional deep sequence models.

    \item We enhance the forecasting model with a unique EMD-based denoising mechanism and an empirical rule-based outlier detection system. This introduces a more efficient way to handle the complexity and variability of internet traffic patterns.

    \item We incorporate a KNN-based outlier mitigation strategy, which uses the average of the nearest $K$ data points to replace identified outliers. This novel approach provides a more effective way to manage and predict network traffic volume demand.

    \item We conduct comprehensive testing and benchmarking of the proposed model using real-world traffic data collected from an ISP provider. The comparative analysis demonstrates the superior performance of our model against other existing state-of-the-art methods.

    \item We make a significant contribution to the field of deep learning and signal processing by proposing a unique approach to internet traffic volume prediction. The proposed model can be adapted for predicting a wide range of time-series trends in various fields like energy, traffic flow, and finance.

\end{enumerate}

This paper is organized as follows. Section \ref{literature} describes the literature review of current traffic prediction using machine learning models. Section \ref{method} presents our proposed methodology. Section \ref{experiment detail} and \ref{Results and Discussion} summarizes the experimentation configuration and discusses results for comparative analysis, respectively. Finally, section \ref{Conclusion} concludes our paper and sheds light on future research directions.

%% file: literature.tex
\section{Literature Review}
\label{literature}
Researchers have recently investigated and proposed various methods for predicting internet traffic. Generally, the three types of network flow forecasts are linear models, nonlinear models, and hybrid models. 

The foundation of a linear traffic forecasting model is the fitting of the polynomial that may capture the trends in historical network traffic data, which are then used to forecast future values. To make the polynomials more closely match the actual internet traffic amount, these linear models must establish various parameters to restrict them. Linear models are extensively used for short-term forecasting and are comparatively fast. As real-world internet traffic is nonlinear, periodic, and has random characteristics, linear models are inappropriate for medium, or long-term forecasting \cite{liter1}. For example, self-similar characteristics and long-range dependence (LRD) of real internet traffic pose challenges to accurately modeling it using AutoRegressive Moving Average (ARMA) methods \cite{yang2021network}. Therefore, various nonlinear prediction models have been proposed to handle complex internet traffic prediction tasks. For example, Fractional Autoregressive Integrated Moving Average (FARIMA) and Seasonal Autoregressive Integrated Moving Average (SARIMA) possess the capability of capturing both short-range dependent (SRD) and LRD characteristics\cite{yang2021farima}\cite{sarima}. All of the aforementioned linear and nonlinear models are individual prediction models. Every prediction model has flaws and issues of its own. As a result, numerous academics have suggested some hybrid prediction models \cite{liter11} where more than two prediction methods are included in the combination forecasting model. These outcomes include a linear model combined nonlinear model, a weight mixture model, decomposition combined model. For example, a new hybrid model that combines two different types of models such as FARIMA or FARIMA/GARCH (Generalized Autoregressive Conditional Heteroskedasticity) and neural network has been proposed in \cite{liter2}. It can handle both long-term and short-term correlations in network traffic. But this method is computationally inefficient. By incorporating FARIMA with alpha stable distribution, researchers have been able to predict wifi traffic with high accuracy and provide Quality of Service (QoS) \cite{sheng2020alpha}. However, this approach is complex and falls short as it fails to reconcile conflicting model properties, and it does not perform well in non-stationary cases that are prevalent in real-world network traffic \cite{christian2021network}. 

Compared to statistical models performance in dealing with non-linear traffic data, neural network based nonlinear models are very popular for handling complex traffic data. For example, Echo State Networks (ESN) \cite{liter4}, Fuzzy Neural Networks \cite{hou2018fuzzy}, and Radial Basis Function Neural Networks \cite{liter5} have been used for traffic prediction. The self-organization and self-learning capabilities of neural networks are strong, and they can effectively explain the nonlinear properties of data. The Echo State Network is a desirable option for network traffic prediction due to its robust nonlinear capabilities and efficient short-term memory capacity. Nonetheless, the generation of the ESN's reservoir is either random or specific, and once created, it remains fixed and cannot be adjusted \cite{8936255}. As a result, the ESN with a fixed reservoir often exhibits sub-optimal performance when dealing with diverse traffic data. To mitigate this limitation, researchers have proposed an ESN with an adaptive reservoir (ESN-AR) \cite{liter4}. However, the use of a Generative Adversarial Network (GAN) to adjust the reservoir leads to computational complexity in the overall process. Although utilizing a loop reservoir in ESN architecture results in a reduction of computational complexity \cite{zhou2022time}, the accuracy of ESN is contingent upon the size of the reservoir, as larger reservoirs may lead to overfitting, thereby limiting the neural network's capacity significantly. Researchers have proposed Fuzzy Neural Network (FNN) models with backpropagation (BP) to address the dynamic mapping problems in the case of internet traffic forecasting \cite{li2019network}. However, BP has certain limitations such as slow learning speed and the tendency to become trapped in local minima \cite{li2006algorithm}. Although FNN-based networks achieve greater accuracy, their high computing demand limits their scalability and simplicity, thereby restricting their utility in production environments. The conventional Radial Basis Function (RBF) networks exhibit limitations such as slow convergence and the possibility to stuck in local optima due to their use of Gradient descent for parameter optimization. Furthermore, this approach leads to slower search speeds and extensive computational requirements, necessitating substantial memory usage. Consequently, the obtained parameters may not be optimal, leading to the limited application of traditional RBF in network traffic prediction. 

\input{literture_table}

The basic principle of the neural network lacks a logical mathematical justification and statistical coherence because it is a "black box" concept. The number of input nodes, output nodes, network layers, nodes in each layer, and other setting methods in the training process, which are relied on experience to establish these hyper-parameters, lack a clear theoretical foundation. Many internet traffic prediction methods based on deep learning have been developed recently due to the advancement of deep learning theory. Long short-term memory (LSTM) \cite{liter7}, deep belief network (DBN), convolutional neural network (CNN) \cite{liter8}, stacked autoencoders (SAE)\cite{wang2018network}, Recurrent Neural Network (RNN) \cite{liter10}, and others are examples of these models. LSTM networks exhibit the ability to retain information in their long-term memory, enabling them to identify patterns of indeterminate lengths. Furthermore, these neural networks can circumvent the vanishing gradient problem that arises with Recurrent Neural Networks (RNNs). However, proposed stacked LSTM networks for anomaly detection in time series\cite{9146846} can be computationally intensive. Other studies have proposed the use of the firefly algorithm with LSTM, introducing the IFA-LSTM method to enhance accuracy \cite{liter7}. Nonetheless, LSTM's learning speed is relatively sluggish when faced with a significant volume of network traffic data. Researchers have also examined the performance of DBN across various Artificial Neural Network architectures. While the DBN exhibits impressive results, the investigation reveals that its effectiveness is heavily reliant on the selection of an optimal number of hidden neurons within each layer \cite{narejo2018application}. Studies observed that beyond a certain point, there is no noticeable correlation between performance and the number of hidden layers, posing a challenge in identifying the ideal number of neurons. Coupled with the inherent complexity of Deep Neural Networks, this feature makes it computationally intensive in real-world applications. On the other hand, due to the growth of smart mobile devices and edge computing devices in the Internet of Things (IoT) has resulted in an ultra-dense network environment with dynamic network topology \cite{li2020traffic}. Consequently, classifying network flow data with high volumes, velocities, varieties, and veracity has become a challenge. To tackle this problem, a new approach was suggested by researchers that involves using multiple Bayesian auto-encoders stacked together to learn the complex relationships among the multi-source network flows \cite{improved-sae}. However, the asymmetry in network traffic can result in suboptimal classification accuracy when using this kind of deep-learning model. 

In recent years, deep learning models have emerged as a powerful approach for time-series prediction due to their ability to automatically learn complex features from the input data. However, deep learning models face challenges in handling the high dimensionality and non-stationarity of time-series data. To address these challenges, researchers have integrated signal decomposition techniques and optimization algorithms with deep learning models to improve their accuracy and efficiency. Signal decomposition techniques such as Fourier analysis, wavelet transforms, and singular spectrum analysis have been used to break down the time-series signal into its constituent parts, such as trend, seasonality, and noise. The extracted features from signal decomposition techniques have then been used as input to deep learning models, leading to improved prediction accuracy. For example, a hybrid model combining Modified Ensemble Empirical Mode Decomposition (MEEMD) and Quantum Neural Network (QNN) has been proposed in \cite{huang2018backbone} and it shown a superior outcomes in capturing characteristics of LRD and SRD, while concurrently reducing computational complexity. Similarly, optimization algorithms such as stochastic gradient descent, Adam, and RMSprop have been used to optimize the parameters of deep learning models, leading to more accurate predictions. For example, researchers suggested various enhancements to the conventional RBF neural network through the implementation of advanced algorithms such as the improved gravitation search algorithm \cite{wei2017network} and particle swarm optimization algorithm \cite{liter6}. The Adaptive Quantum Particle Swarm Optimization Algorithm (AQPSO) has demonstrated improved results in \cite{zhou2022self}. Nevertheless, due to the excessive number of parameters involved in RBF optimization, the computation scale and training time increase considerably, impeding the convergence rate. As a result, the applicability of RBF in small networking devices with limited computational capacity becomes constrained.

\begin{figure*}[h]
\label{fig:method}
    \centering
    \includegraphics[height=9cm,width=17cm]{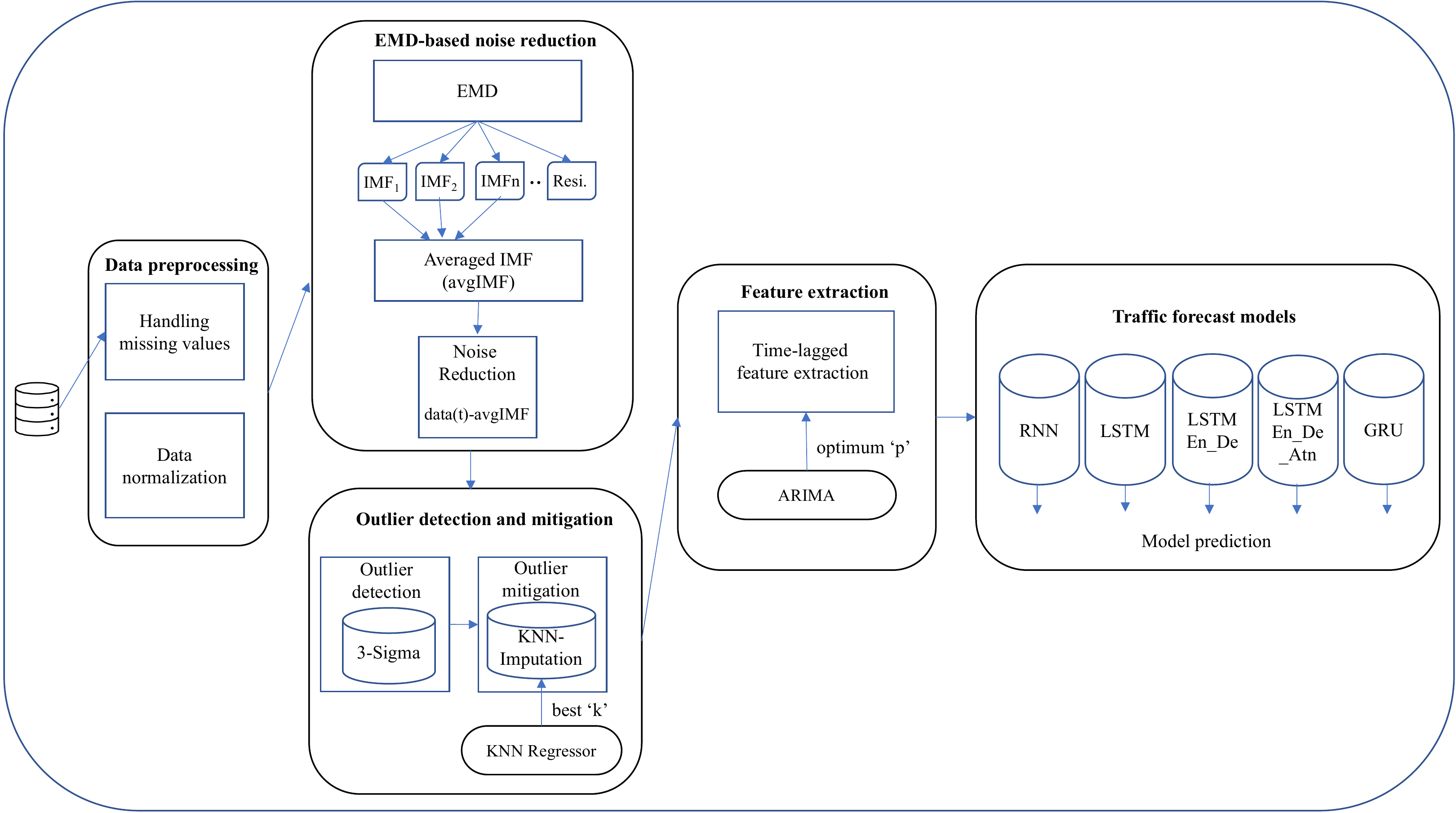}
    \caption{High-level framework of our proposed methodology.}
    \label{fig:methodology}
\end{figure*}

According to the comparative summary of existing works in Table 1, we noticed different types of prediction models. There are still some issues with choosing the prediction model and its parameters, carrying out a reasonable fusion of the prediction results, and selecting the right decomposition or optimization algorithm, despite the fact that many research results have shown that the combination prediction model has achieved better performance than the single prediction model. If these issues cannot be adequately resolved, it may result in the hybrid prediction model not outperforming the single prediction model in terms of performance metrics despite the computational complexity being increased. Moreover, we noticed a lack of investigation in noise reduction and outlier mitigation in traffic prediction. Both noise reduction and treatment of outlier data points are crucial pre-processing methods for time-series prediction, but they focus on various sorts of data imperfections. While outlier data point handling strategies try to find and fix data points that dramatically depart from the expected pattern, noise reduction approaches strive to eliminate random shifts in the time-series data. The specific attributes of the time-series data and the objectives of the prediction task determine which techniques should be used. Therefore, in this work, we proposed a novel traffic prediction framework called DEK-Forecaster integrating deep (D) sequence model with Empirical Mode Decomposition (E) based traffic denoising and K-Nearest Neighbour (K) based outlier mitigation and performed a comprehensive experiment with a real-world traffic dataset.

%% file: literture_table.tex
\begin{table*}
\centering
\caption{Comparative summary among existing works. }
\begin{tabular}{|m{.4cm}|m{2cm}|m{2cm}|m{2.1cm}|m{2.2cm}|m{3cm}|m{3cm}|} 
\hline
\textbf{Ref.} & \textbf{Decomposition }                                                                         & \textbf{Noise Reduction}                                                                                                                                                     & \textbf{Outlier Detection } & \textbf{Prediction Model}                                 & \textbf{Benefits}                                                                         & \textbf{Drawbacks}                                                                                                                                                \\ 
\hline

\cite{liter4}               & N/A                                                                                             & Local Preserving Projection(LPP)                                                                                                                                             & N/A                         & ESN with double loop reservoir                             & Accuracy improved after denoising the data using LPP                                  & The performance is critically influenced by the identification of the optimal value of the "Window size" hyperparameter                                                             \\ 
\hline
    
\cite{hou2018fuzzy}               & N/A                                                                                             & N/A                                                                                                                                                                          & N/A                         & ARIMA–SA–BPNN                                             & Improved accuracy by optimizing network weights with SA(Simulated Annealing) algorithm                    &The analysis was conducted using only two datasets                                                                                                                                    \\

\hline

\cite{li2019network}               & Glowworm Swarm Algorithm                                                           & Auto-correlation method with average displacement & N/A                         & BP neural network                                         &   The use of the Glowworm Swarm Algorithm resulted in a higher level of accuracy when compared to a standard Backpropagation Neural Network(BPNN).                                          & Incorporating the auto-correlation technique along with the calculation of average displacement increases computational complexity.                                                                   \\ 
\hline

\cite{liter3}               & N/A                                                                                             & N/A                                                                                                                                                                          & N/A                         & Threshold-based FARIMA                              & The Mean Squared Error (MSE) showed a 72\% improvement compared to the conventional FARIMA method       & The effectiveness of filtering depends on filter order, difference coefficient, and test data quantity                                   \\ 
\hline
\cite{huang2018backbone}               & Modified Ensemble Empirical Mode Decomposition (MEEMD)& Modified Ensemble Empirical Mode Decomposition (MEEMD)                                                                           & N/A                         & Quantum Neural Network (QNN)                              & Better identification of long-range dependence (LRD) and short-range dependence (SRD) & Computing resource requirements of the method pose challenges for low-resource devices                                                                                                             \\ 
\hline

\cite{littable1}           & N/A                                                                                             & Autoencoders                                                                                                                                                                 & N/A                         & SARIMA, Hybrid of LSTM and Autoencoder                         & Achieved a 50\% reduction in false positive rate                                                   & The testing was performed exclusively on a dataset that followed a particular distribution 
\\
\hline
\cite{liter8}              & Ensemble Empirical Mode Decomposition                                                           & Ensemble Empirical Mode Decomposition                                                                                                                                        &             N/A                & Three-dimensional convolutional neural network (3D CNN) & 3D CNN outperforms 2D CNN in extracting spatial-temporal features              &       Does not take into account medium-term or long-term traffic prediction                                                                                                                                                        \\ 
\hline
\cite{narejo2018application}               & NA                                                                                              & N/A                                                                                                                                                                          & N/A                         & Deep Belief Network (DBN)                                 &     Pretraining a Deep Belief Network (DBN) using unsupervised learning techniques enhanced its performance                                                                                  & While accurately selecting hyperparameters is crucial for achieving the desired outcome, no hyperparameter tuning method has been suggested.  \\
\hline
\end{tabular}
\end{table*}

%% file: methodology.tex
\section{Our Proposed Forecasting Methodology}
\label{method}
This section briefly explains our proposed traffic prediction methodology depicted in Fig. \ref{fig:methodology}. 
\begin{algorithm}
\label{algo:weighted_voting}
\SetAlgoLined
\SetKw{KwBy}{by}
\SetKwInOut{Input}{Input}\SetKwInOut{Output}{Output}
\caption{Proposed traffic prediction algorithm with noise filtering and outlier handling.}
\textbf{Input: } Traffic data, $y(t)$\\
Apply EMD on actual traffic data, $y(t)$, and obtain $IMF_n(t), \text{ for n} = 1,2,...,I$\\
Calculate average IMF, avgIMF(t) = $\dfrac{\sum_{i=0}^{I} IMF_i(t)}{I}$\\
Remove noise from original traffic, $y(t) = y(t) - avgIMF(t)$\\
Calculate average traffic, $AVG_{y(t)}$=$\dfrac{\sum_{i=0}^{N} y(i)}{N}$\\
Calculate standard deviation, $SD_{y(t)}= \sqrt{\dfrac{\sum_{i=0}^{N}(y(i)-AVG_{y(t)})^2}{N}}$\\
Traffic upper limit, $u_{lim} = AVG_{y(t)} + 3*SD_{y(t)}$\\
Traffic lower limit, $l_{lim} = AVG_{y(t)} - 3*SD_{y(t)}$\\
k\_list = $[2\text{ to }24]$\\
MIN\_RMSE = INF\\
BEST\_K = 0\\
\ForEach{$K$ in $k\_list$}
{
    apply KNN regressor on $y(t)$ for next-step prediction\\
    \If{$prediction\_RMSE < MIN\_RMSE$}{
    BEST\_K = k 
    }
}
\For {$i=1,2,\ldots,N$}{
\If{$y(i)>u_{lim}\mid y(i)<l_{lim}$}
{
y(i) = Average of BEST\_K nearest data points based on Euclidean distance
}
}
\textbf{Output: }Noise free and outlier adjusted traffic, y(t)
\end{algorithm}

\subsection{Empirical Mode Decomposition Based Noise Reduction}
\label{label:noise_reduction}
Empirical Mode Decomposition (EMD) is a technique to extract several components from a signal assuming every signal comprises of sub-components. This approach is also known as Hilbert–Huang transforms (HHT) and is extensively used for time-frequency analysis of non-stationary and non-linear time series data. EMD decomposed an original signal into several zero mean and quasi-periodic components called Intrinsic Mode Functions (IMF) alongside a residue element representing the trend as shown in Eq.\ref{eq:emd} where each $h_i(t)$ stands for the ith IMF, $r(t)$ is the residual component, and $y(t)$ is the original value of the data.
\begin{equation}
\label{eq:emd}
    y(t) = \sum_{i}^{n}h_i(t) + r(t)
\end{equation}
Real-world internet traffic has random, non-stationary characteristics influenced by various external and internal factors related to ISP companies. These factors can be categorized as geographic factors, economic factors, ISP new service, service decommission factors, weather, time, day, season, special event, etc. Due to these factors, the ISP traffic is composed of many individual components, and EMD can be helpful for better analysis and forecasting of internet traffic. Noise filtering or reduction or signal denoising is a process of removing noise from time-series data. Any time series may consist of three systematic elements: level, trend, and seasonality, and one non-systematic element, noise. The noise reduction approach for better learning and forecasting by the machine learning model should minimize noise elements in time series. Among different noise filtering approaches, EMD based denoising technique has been applied extensively in different areas. Classic EMD-based denoising techniques choose a particular IMF component to eliminate noise elements from the signal, but there is no formal logic for selecting an IMF from decomposed components. Moreover, choosing one IMF for denoising is difficult as the number of IMF depends on the original signal. Therefore, a new EMD-based noise filtering method has been proposed in \cite{method1}. They showed that the average of all IMF (avgIMF) components is normally distributed. The avgIMF corresponds to the maximum signal noise level and has the most white Gaussian noise features of any IMF element. 

The EMD technique is used in this study to remove noise from the internet traffic time series data, which is considered one-dimensional data. The abrupt changes in our traffic forecasting data have been smoothed by removing noise based on steps 3 and 4 in Algorithm 1. We extracted the average of all IMF elements, avgIMF, from our original signal, $y(t)$, to obtain a noise-free traffic data $y_n(t)$. 


\subsection{Empirical Rule and KNN Based Outlier Management}
\label{label:outlier_detection}
Outlier detection is a necessary preprocessing step for real-world internet traffic analysis. The data points significantly different from most of the values are considered outliers. Outliers are characteristically different than noise in the time series. Noise is a random error in the data and needs to be removed entirely from the original signal for a better prediction model. On the contrary, the outliers are the part of the time series that impacts different statistical parameters, such as mean, standard deviation, correlation, etc., of the original signal. Outliers can lead to incorrect future predictions of internet traffic. 

A statistical principle known as the empirical rule, the three-sigma rule or 68-95-99.7 rule, holds that almost all observed data will lie within three standard deviations of the mean with a normal distribution. However, this rule is also applicable for non-normally distributed data where 88.8\% of data fall within the three-sigma interval as opposed to 99.97\% for normal distribution. According to Chebyshev's inequality, 75\% of the data lie inside two standard deviations for a wide range of various probability distributions, while the empirical rule claim 95\% data points within the second standard deviation for normal distribution\cite{method2}. In this work, we set an upper and lower limit for most of the data instances in our original signal. The individual data point is outside the three-standard deviation. We considered them as point outliers. Those point outliers in our dataset have been mitigated by $K$ nearest data points based on the standard KNN-Imputation algorithm. Each outlier point is imputed using the mean value from $K$ nearest neighbors in the training set. The training dataset's members' distances are calculated using a Euclidean distance measure that is NaN aware, which excludes NaN values from the calculation. For optimum $K$ value, we apply KNN-regressor in our traffic dataset where past observations are used to predict the following data points, and this experiment has been conducted for different $K$ values ranging from 2 to 24. The minimum prediction error measured in terms of RMSE (Root Mean Squared Error) is the criteria for best choosing the best $K$ for imputing outlier data points.

\subsection{Time-lagged Feature Extraction}
\label{label:feature extraction}
Generally, the time series prediction task uses previous data samples to predict the following values. We extract the time-lagged feature from our original dataset for training and testing our prediction model in this work. Based on ACF analysis in subsection \ref{acf}, we concluded that our traffic data is non-random. Hence, we considered previous timestamps features for our deep-learning models to predict the following values. We performed a grid search based on the Akaike Information Criterion (AIC) value and a statistical prediction model called ARIMA to determine the optimum number of lagged features. The AIC measures the relative quality of statistical models for a given set of data and predicts prediction error. AIC calculates each model's efficiency in relation to the other predictions given a set of models for the data. As a result, AIC offers a model ranking method. However, we compare ARIMA model prediction performance with various settings of hyperparameters and rank them based on the AIC value. ARIMA model predicts the future value based on the past values of the time series, that is, its own lagged values. The model requires three parameters such as $AR(p)$, $MA(q)$, and $I(d)$, which represent the Autoregressive, Moving Average and differencing order. Among these parameters, the AR term defines the number of lagged features used to forecast the next value. Therefore, we performed a grid search using a different combination of $p$, $q$, and $d$ to perform single-step prediction using the ARIMA model and select the best model by comparing their prediction performance based on our selection criteria. We consider AR term, $p$, from best performing model based on minimum AIC and $p$ indicates the time-lagged feature which gave us better prediction. So, the prediction task in this study is performed as in Eq.\ref{eq:fe}. where $y(t)$ is the traffic volume for the current time step, $y(t-1)$ to $y(t-p)$ represents the previous $p$ data points, and $h$ is the prediction function.
\begin{equation}
\label{eq:fe}
    y(t+1) = h(y(t-1), y(t-2), ...., y(t-p))
\end{equation}

\subsection{Time and Space Complexity Analysis of Proposed Method}
There are several individul module in our proposed model. Among them, time complexity for EMD based noise reduction is $O(n^2)$ where $n$ is the length of the input data. The time complexity of calculating the average IMF signal is $O(i*n)$, where $i$ is the number of IMFs and $n$ is the length of the input data. Finally the noise removal step  involves subtracting the average IMF from the original traffic data, which has a time complexity of $O(n)$. The outlier management module time complexity depends on outlier detection and outlier mitigation. Outlier detection step involves calculating the average and standard deviation of traffic, which both have a time complexity of $O(n)$. Then the outlier mitigation steps requires to execute KNN algorithm and its time complexity is $O(n*log(n)*k)$. Therefore, the overall time complexity of our propsoed model is dominated by the EMD-based noise reduction step, which has a time complexity of $O(n^2)$. Thus, the time complexity of our proposed algorithm can be approximated as $O(n^2)$.

The space complexity of the EMD algorithm is $O(n^2)$, as it involves storing the decomposed IMFs. The space complexity of calculating the average IMF signal is $O(n)$, as it involves storing the average IMF signal. The noise removal step involves modifying the original traffic data in-place, so it does not require any additional space. In case of outlier detection, calculation of average traffic and standard deviation requires storing a single value, so their space complexity is $O(1)$. Then finding traffic upper and lower limit involves storing two values, so the space complexity is O(1). The outlier detection step needs to store a list of $K$ values, so the space complexity is $O(k)$. As we used KNN regressor to find the best $K$, it has a space complexity of $O(n*k)$. Finally, outlier handling involves modifying the original traffic data in-place, so it does not require any additional space. Therefore, the overall space complexity of the algorithm is also dominated by the EMD step, which has a space complexity of $O(n^2)$. Thus, the space complexity of the algorithm can be approximated as $O(n^2)$.

%% file: result.tex
\section{Experiment Detail}
\label{experiment detail}
\subsection{Dataset Description}
In this study, we utilized a dataset comprised of real-world traffic volume for our experiments. We gathered telemetry information by sampling the value of the 'ifOutOctets' counter on a core-facing interface of a provider edge router's SNMP (Simple Network Management Protocol) interface. To calculate the bps (bits per second) value for each interval, we multiplied the difference between observations at both ends of the interval by eight. Each sampling was performed at five-minute intervals. Given that we were working with a 40 Gbs connection, no skips occurred during the sampling duration (the 'ifOutDiscards' remained unchanged).

Our dataset incorporated 29 days' worth of traffic volume data, equating to a total of 8,352 data samples. We allocated 70\% of this data, approximately equivalent to 21 days' worth of traffic data, for the purpose of training our prediction model. The remaining data, covering the subsequent seven days, was utilized for testing our proposed forecasting model.

\subsection{Data Preprocessing}
In this section, we undertake certain data preprocessing tasks before initiating the training of our prediction model.
\subsubsection{Managing Missing Values and Data Normalization}
For our traffic data, we used the most recent valid data entry to replace any missing values. While there are multiple approaches to dealing with missing data in time series analysis, including mean value replacement, linear interpolation, and quadratic interpolation, we found these methods to be unsuitable for our dataset. The linear interpolation method assumes a linear relationship between data points and replaces null values by drawing a straight line between these points. However, given the non-linear nature of our dataset, this approach proved ineffective. The polynomial interpolation method requires predefining the order to replace missing values, which could lead to inaccuracies as it fits the smallest possible degree polynomial through the available data points. Mean value replacement was also unsuitable as real-world traffic measurements can contain outliers and unexpected data points, which can skew the mean. Ultimately, we adopted the forward-filling strategy typically used for time-series data, where the previous value is used to fill in the subsequent missing value. This approach was found to be beneficial for our experiment.

\begin{figure}[h]
    \centering
    \includegraphics[width=8cm,height=2.5cm]{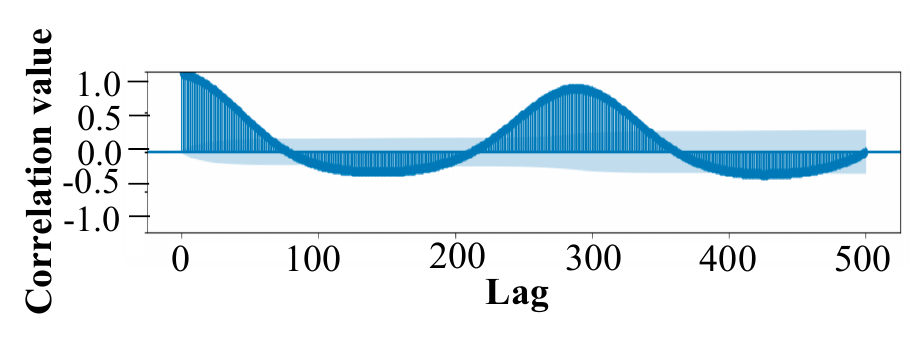}
    \caption{ACF plot of our traffic data.}
    \label{fig:acf}
\end{figure}

\subsubsection{Traffic AutoCorrelation Function (ACF) Analysis}
\label{acf}
An AutoCorrelation Function (ACF) plot serves as a useful tool to evaluate the stationarity and predictability of time-series data. Additionally, data from the ACF plot can inform us about the seasonality and trends within the time-series data. Each bar in an ACF plot represents the strength and direction of correlation between data points at different lags. For data characterized as random, or white noise, the bar values for all lags should be close to zero. In contrast, non-random time-series data will exhibit at least one lag value with strong correlation. This allows us to construct a predictive model utilizing time-lagged features for such non-random time-series data. As depicted in Fig. \ref{fig:acf}, our traffic data demonstrates considerable lags along with high correlations. For instance, the correlation value is near 1 for lags such as 1, 24, etc., which indicates that our traffic data is not random. This observation supports the decision to incorporate time-lagged features into our regression model, enhancing its accuracy.

\subsection{Evaluation Metrics}
\label{metric}
We used Root Mean Squared Error (RMSE), Mean Absolute Error (MAE), and Mean Average Percentage Error (MAPE)to estimate the performance of our traffic forecasting models.  RMSE is widely used in time-series prediction because it takes into account the magnitude of the errors. This means that larger errors are penalized more heavily than smaller errors, which is useful when the size of the errors matters. MAE is another well-known metric to evaluate the average magnitude of the errors. Unlike RMSE, MAE treats all errors equally, which can be useful to evaluate the model's performance in terms of the absolute deviation of the predictions from the true values. If the scale of the target variable is large then MAPE can be used to evaluate the accuracy of the model's predictions in percentage terms. We define our performance metrics mathematically as follow, where $y_i$ and $\hat{y}_i$ are predicted, and the original value, respectively, and  $n$ is the total number of a test instance.

\begin{equation}
RMSE = \sqrt{\frac{\sum_{i=1}^{n}(y_i - \hat{y}_i)^2}{n}}
\end{equation}

\begin{equation}
MAE = \frac{1}{n} \sum_{i=1}^{n} \left| y_i - \hat{y}_i \right|
\end{equation}

\begin{equation}
MAPE = \frac{100\%}{n} \sum_{i=1}^{n} \left|\frac{y_i - \hat{y}_i}{y_i}\right|
\end{equation}

\subsubsection{Software and Hardware Preliminaries}
\label{software}
We used Python and the machine learning library Tensorflow-Keras\cite{chollet2015keras} to conduct the experiments.  Our computer has the configuration of Intel (R) i3-8130U CPU@2.20GHz, 8GB memory, and a 64-bit Windows operating system. 

\section{Result Analysis and Discussion}
\label{Results and Discussion}
This section provides an overview of the performance of our proposed traffic model. We first delve into the results from the optimal feature selection using ARIMA in subsection \ref{feature}. We then turn our attention to the performance of our model that integrates outlier detection and mitigation in subsection \ref{outlier identify and mitigation}. Lastly, we examine the efficacy of the traffic denoising component within our proposed traffic model in subsection \ref{traffic denoise}.

\subsection{Optimum Feature Selection for Prediction Model}
\label{feature}
We identify the optimal time-lagged feature sets for our deep learning prediction model based on ARIMA model performance in single-step prediction. The AR term in ARIMA refers to the number of lagged values of the dependent variable (i.e., the time series data) that are used to predict future values. The AR term is denoted by $p$ and it indicates the order of autoregression. For example, ARIMA$(p, d, q)$, where $p$ represents the order of the AR term. Table \ref{tab:aic} summarizes five best-performing ARIMA model hyper-parameters and their corresponding AIC value. We considered AR and MA term ranges from 2 to 24 for finding the best combination of model parameters so that we can select the optimum AR term that represents the most relevant time lag for capturing the temporal dependencies in the traffic data. This time-lagged feature was then incorporated into our deep learning prediction model to enhance its performance. According to our experimental result, the ARIMA model with the lowest AIC value, which was ARIMA (13, 1, 16) with an AIC of 3782.588307, is the best-performing model for our analysis. Therefore, we selected 13 time-lagged features to train our deep sequence model for traffic prediction. 

\begin{table}[!htbp]
\caption{Top ten best-performing model parameter configuration}
\label{tab:aic}
\centering
\begin{tabular}{lcc} 
\hline
SL. & ~ ~(p, d, q)~ ~ & ~ ~ ~AIC~ ~~  \\ 
\hline
1  & (13, 1, 16)     & 3782.588307      \\
2  & (21, 1, 2)     & 3782.751381   \\
3  & (14,1,17)      & 3783.706900   \\
4  & (13, 1, 18)     & 3785.061129   \\
5  & (21, 1, 3)     & 3785.332434   \\
6  & (22, 1, 3)     & 3786.325772      \\
7  & (16, 1, 17)     & 3786.547150   \\
8  & (17,1,17)      & 3786.761965   \\
9  & (22, 1, 4)     & 3786.964070   \\
10  & (20, 1, 2)     & 3787.326614   \\
\hline
\end{tabular}
\end{table}

\subsection{Performance Analysis of Prediction Model with Outlier Management}
\label{outlier identify and mitigation}
In this section, we discuss the characteristics of outliers identified by our outlier detection module. Then we analyze the impact of outlier management on traffic prediction tasks compared with several baseline models. 

\begin{figure}[h]
    \centering
    \includegraphics[height=3cm, width = 8cm]{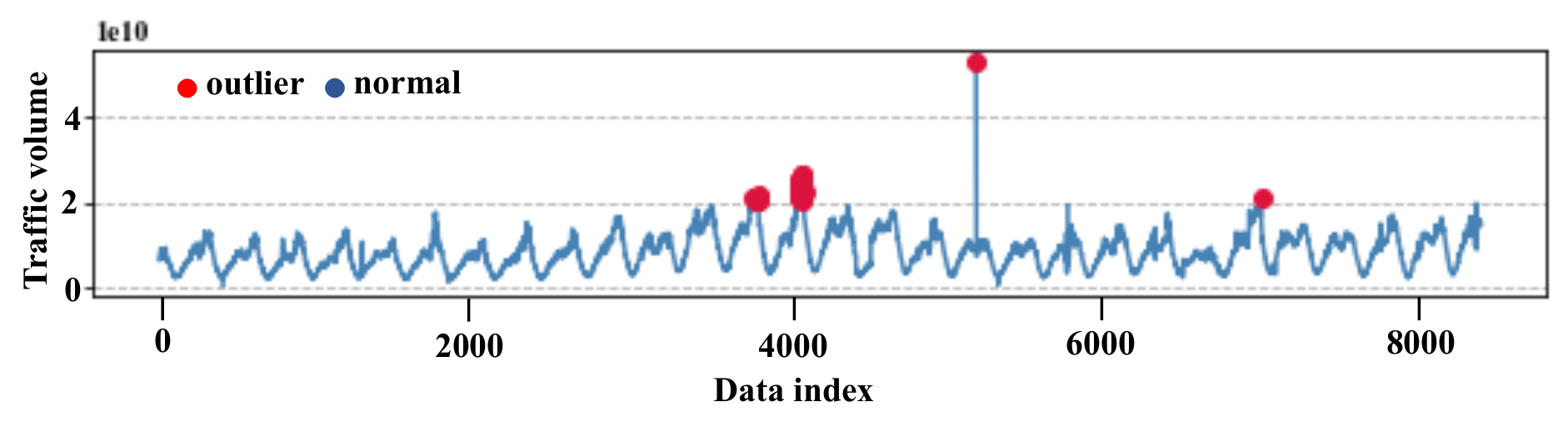}
 \caption{Outlier points identified using empirical rule.}
    \label{fig:outlier}
\end{figure}

\begin{figure}[h]
    \centering
    \includegraphics[height=3cm, width = 8cm]{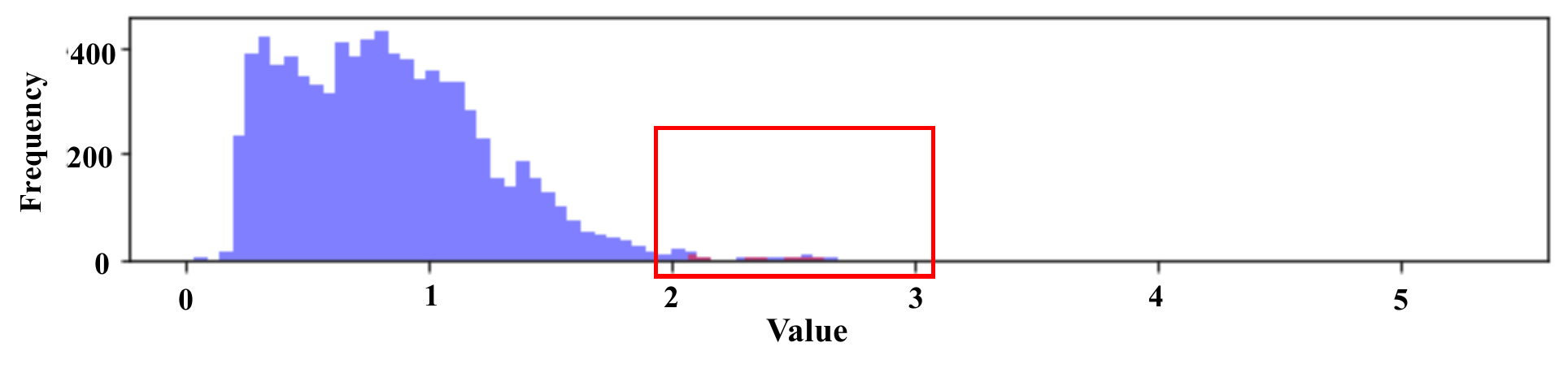}
    \caption{Distribution of data with outlier points highlighted.}
    \label{fig:outlier_dist}
\end{figure}

Based on our experimental results, there are 43 outlier data samples in our traffic dataset, which lie outside three standard deviations from the mean value, depicted in Fig.  \ref{fig:outlier}. We analyze some statistical properties of outliers samples based on their distribution in Fig. \ref{fig:outlier_dist}. It is a right-skewed histogram where the data is clustered towards the left side of the histogram and extends further to the right. This type of distribution is also called a positive skew. In this diagram, we identified the outlier points on the right side of the right-skewed histogram and marked that portion using a red rectangle. A right-skewed histogram with an outlier on the right side indicates that at least one data point has an extremely high value relative to the rest of the data, and Fig. \ref{fig:outlier} depicts several data points which are relatively very high. These outliers can have a significant impact on the overall distribution of the data, as they can affect the mean and median of the dataset. In general, it is important to identify and handle outliers in a dataset, especially if they are affecting the distribution of the data. Therefore, we treated those outlier points before using them to train our prediction model. We applied KNN imputation to handle these outlier data points, and the basic idea behind KNN imputation is to replace the outlier value with the average of the $K$ nearest neighbors. One of the core challenges of this approach is to find the best value for $K$. A smaller value of $K$ will result in a stricter imputation estimate, focusing on a local perspective of the domain. On the other hand, a larger value of $K$ will result in a more general estimate, focusing on a global perspective of the domain. A small $K$ value may result in over-fitting, where the imputed values are too similar to the original data and may not be accurate. A large $K$ value may result in under-fitting, where the imputed values are too different from the original data and may not be representative of the underlying distribution. Therefore, we used the KNN regressor to determine the optimal value of $K$. A range of $K$ values has been considered in our experiment and reported their corresponding RMSE in Table \ref{tab:rmse for kvalues} for predicting the next data value. The lower RMSE indicates a strong relationship among $K$ previous values to forecast the next value in the series. And based on this, we replaced the outlier data samples with the average of the $K$ nearest neighbors traffic volume data. According to our experimental results in Table \ref{tab:rmse for kvalues}, $K$ value 11 gave us the lowest RMSE of 1.78 Gbps. Based on this, we considered the average of 11 previous data points to replace corresponding outlier data.

\begin{figure*}[h]
    \centering
    \includegraphics[width=16cm,height=5.5cm]{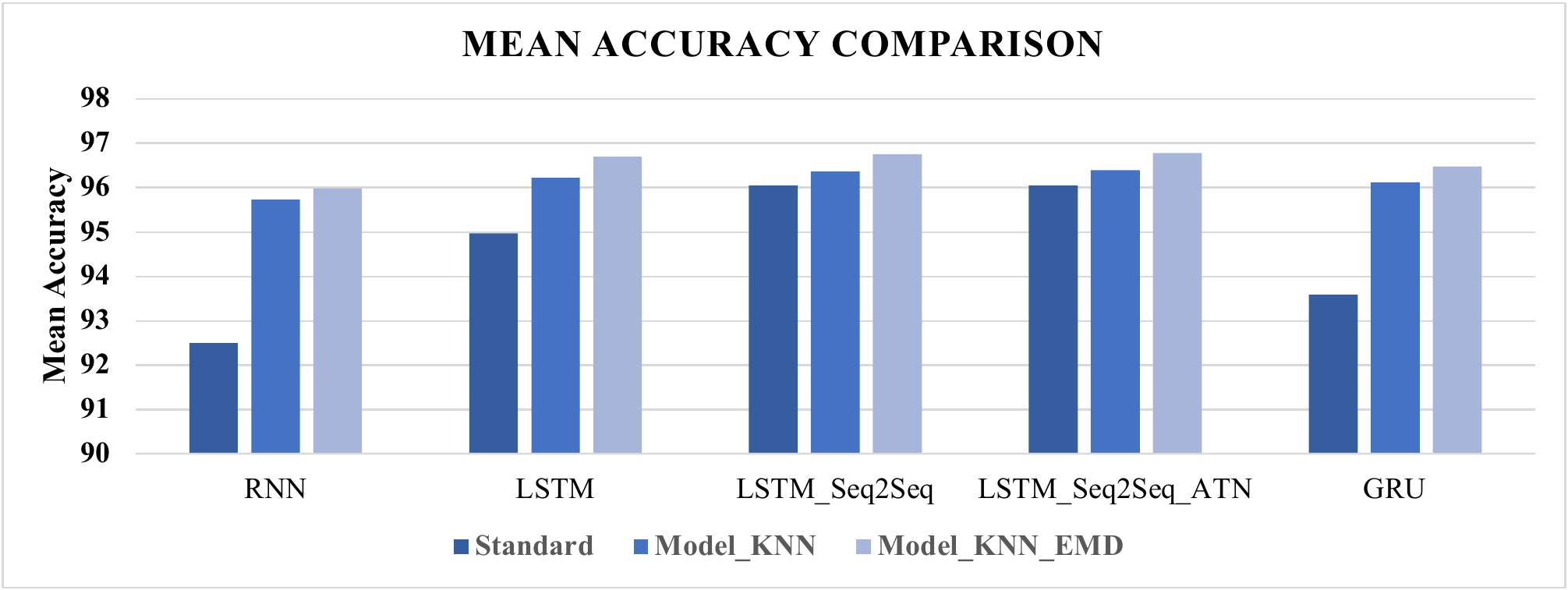}
    \caption{Prediction accuracy comparison between conventional model and proposed KNN-EMD integrated model.}
    \label{fig:acc_standard_knn_emd}
\end{figure*}

\begin{figure*}[h]
    \centering
    \includegraphics[width=17cm,height=5cm]{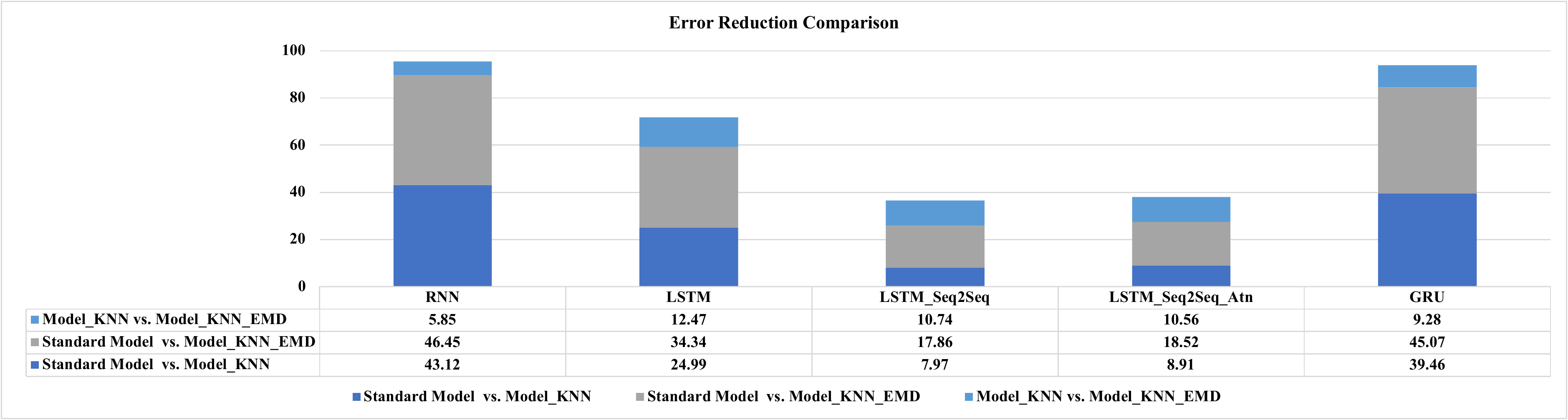}
    \caption{A comparison of prediction error reduction (\%) by best-performing techniques with standard deep learning model and another approach.}
    \label{fig:outliernoise_vs_standard}
\end{figure*}

After treating those outlier data samples in our traffic dataset, we compare our KNN-integrated traffic prediction model performance with several baseline deep learning models. The baseline conventional deep learning models used in the analysis are RNN, LSTM, LSTM\_Seq2Seq, LSTM\_Seq2Seq\_ATN, and GRU. These models were evaluated based on the performance metrics such as RMSE, MAE, and MAPE, and all results are summarized in Table \label{tab:all performance}. Our proposed deep learning model integrated with KNN-based outlier detection outperforms the conventional model. As depicted in Fig. \ref{fig:outliernoise_vs_standard}, the prediction error dropped significantly for KNN-integrated models compared to the standard deep learning models. For example, in the case of RNN, the prediction error has been reduced from 7.51\% to 4.27\% when outliers were handled, and it is approximately 43\% less error compared to traditional RNN. Similarly, LSTM\_KNN, LSTM\_Seq2Seq\_KNN, LSTM\_Seq2Seq\_Atn\_KNN, and GRU\_KNN gave better prediction accuracy compared to their corresponding traditional model. From Fig. \ref{fig:outliernoise_vs_standard}, we noticed approximately 24\%, 8\%, 9\%, and 40\% less error we achieved respectively for LSTM\_KNN, LSTM\_Seq2Seq\_KNN, LSTM\_Seq2Seq\_Atn\_KNN, and GRU\_KNN by managing outliers before using them to train our model. Consequently, we noticed a significant accuracy improvement in KNN integrated models compared to the conventional model in Fig. \ref{fig:acc_standard_knn_emd}. Overall, incorporating KNN-based outlier mitigation in conventional deep learning models helped improve prediction accuracy by reducing the influence of outliers or extreme values in the input data. Outliers can be problematic in time-series prediction models, as they can introduce noise and bias into the model and cause it to make inaccurate predictions. By removing or reducing the influence of outliers, the model is able to learn more effectively from the remaining less noisy data, and our experimental results also indicate the adverse effect of outlier data points in model performance.

In our study, we evaluated the performance of various models for internet traffic forecasting using three error metrics: Root Mean Square Error (RMSE), Mean Absolute Error (MAE), and Mean Absolute Percentage Error (MAPE). As shown in Table \ref{tab:all performance}, the LSTM\_Seq2Seq\_ATN\_KNN model achieves the lowest RMSE (0.59), MAE (0.31), and MAPE (3.60) values, indicating the best overall performance among the evaluated models. The consistent pattern of low error values across all three metrics suggests that this model provides accurate predictions in both relative and absolute terms. The other models, such as LSTM\_Seq2Seq\_KNN and GRU\_KNN, also exhibit a similar trend, with lower MAPE values corresponding to lower RMSE and MAE values. These results demonstrate the effectiveness of our chosen models in predicting internet traffic, with the LSTM\_Seq2Seq\_ATN\_KNN model outperforming the others. The consistency in the error metrics suggests that our model is performing well and is accurately capturing the underlying patterns in the data. A low MAPE means that the model's predictions are close to the actual values in relative terms, while low RMSE and MSE values imply that the model's predictions are also close to the actual values in absolute terms.

\begin{table}
\centering
\caption{KNN-Regressor prediction error (RMSE) for different $K$ values. } 
\label{tab:rmse for kvalues}
\begin{tabular}{|c|c|} 
\hline
\textbf{K-value} & \textbf{RMSE}  \\ 
\hline
2                & 1856644563     \\ 
\hline
3                & 1828291277     \\ 
\hline
4                & 1816389491     \\ 
\hline
5                & 1826662669     \\ 
\hline
6                & 1811743008     \\ 
\hline
7                & 1815585493     \\ 
\hline
8                & 1817635025     \\ 
\hline
9                & 1810554111     \\ 
\hline
10               & 1797064368     \\ 
\hline
11               & 1796783992     \\ 
\hline
12               & 1804298868     \\ 
\hline
13               & 1809809913     \\ 
\hline
14               & 1814868475     \\ 
\hline
15               & 1818749549     \\ 
\hline
16               & 1831109596     \\ 
\hline
17               & 1841230237     \\ 
\hline
18               & 1856738081     \\ 
\hline
19               & 1866925893     \\ 
\hline
20               & 1881460309     \\ 
\hline
21               & 1892611532     \\ 
\hline
22               & 1902113203     \\ 
\hline
23               & 1914844430     \\ 
\hline
24               & 1922448158     \\
\hline
\end{tabular}
\end{table}

\begin{table}
\centering
\caption{Signal-to-Noise(SNR) ratio comparision.}
\label{tab:snr}
\begin{tabular}{|l|l|} 
\hline
                    & Signal-to-Noise (SNR) Ratio  \\ 
\hline
Noisy signal SNR    & -7.05 dB                     \\ 
\hline
Denoised signal SNR & 21.47 dB                     \\
\hline
\end{tabular}
\end{table}

\begin{table}
\centering
\caption{Performance Summary of Proposed Prediction Models}
\label{tab:all performance}
\begin{tabular}{l|c|c|c} 
\hline

\multicolumn{4}{c}{\bf{Baseline Model}}                                                \\ 
\hline\hline
Model                        & RMSE          & MAE                     & MAPE                          \\
\hline
RNN                          & 1.49 & 0.84           & 7.51                       \\
LSTM                         & 1.22 & 0.62            & 5.03                        \\
LSTM\_Seq2Seq                & 0.58  & 0.34           & 3.94                        \\
LSTM\_Seq2Seq\_ATN            & 0.58  & 0.34           & 3.95                       \\
GRU                          & 1.47 & 0.75           & 6.41                       \\ 
\hline\hline
\multicolumn{4}{c}{\bf{Proposed Model}}                                                                          \\ 
\hline\hline
Model                        & RMSE          & MAE                     & MAPE                         \\
\hline
RNN\_KNN                     & 0.76  & 0.49          & 4.27                       \\
LSTM\_KNN                    & 0.64  & 0.33        & 3.77                       \\
LSTM\_Seq2Seq\_KNN            & 0.60 & 0.32         & 3.63                       \\
LSTM\_Seq2Seq\_ATN\_KNN       & 0.59  & 0.31          & 3.60                       \\
GRU\_KNN                     & 0.63  & 0.33      & 3.88                       \\ 
\hline\hline
\multicolumn{4}{c}{\bf{Proposed Model}}                                                                          \\ 
\hline\hline
Model                        & RMSE          & MAE                     & MAPE                           \\
\hline
RNN\_KNN\_EMD                & 0.60 & 0.345           & 4.02                       \\
LSTM\_KNN\_EMD               & 0.57  & 0.30            & 3.30                        \\
LSTM\_Seq2Seq\_KNN\_EMD      & 0.55 & 0.29           & 3.24                        \\
LSTM\_Seq2Seq\_ATN\_KNN\_EMD & 0.54  & 0.28           & 3.22                       \\
GRU\_KNN\_EMD                & 0.57 & 0.31          & 3.52                       \\
\hline\hline
\end{tabular}
\end{table}
\begin{figure}[h]
    \centering
    \includegraphics[width=8cm,height=3cm]{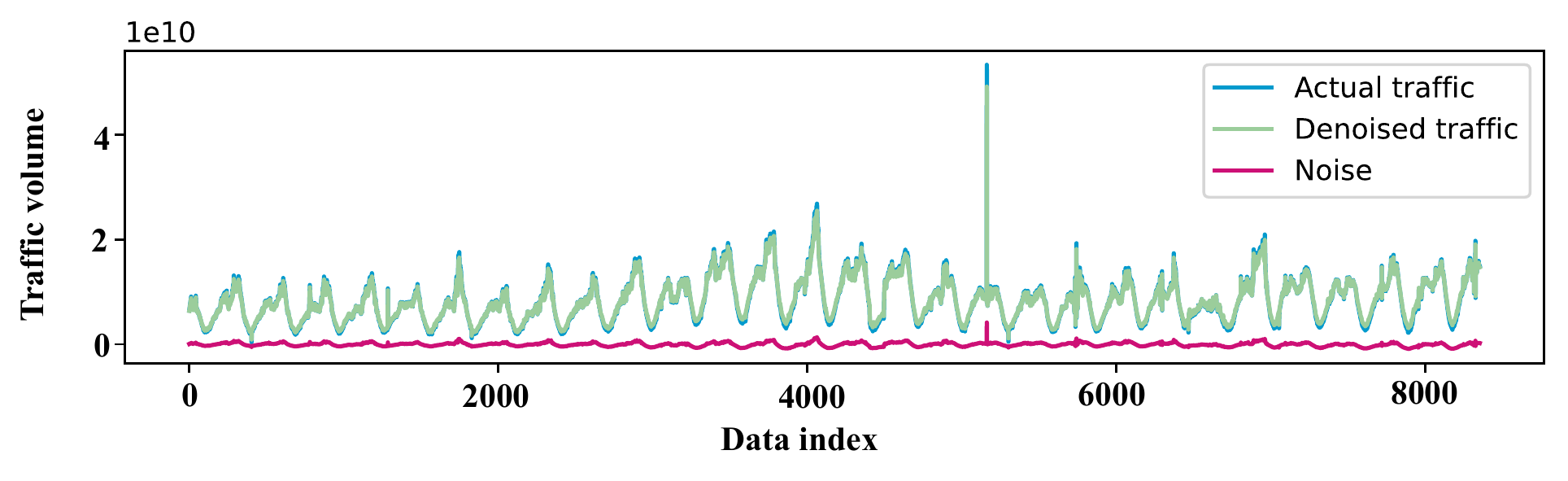}
    \caption{Denoised traffic data.}
    \label{fig:denoise}
\end{figure}

\begin{figure*}[h]
    \centering
    \includegraphics[width=15cm,height=4cm]{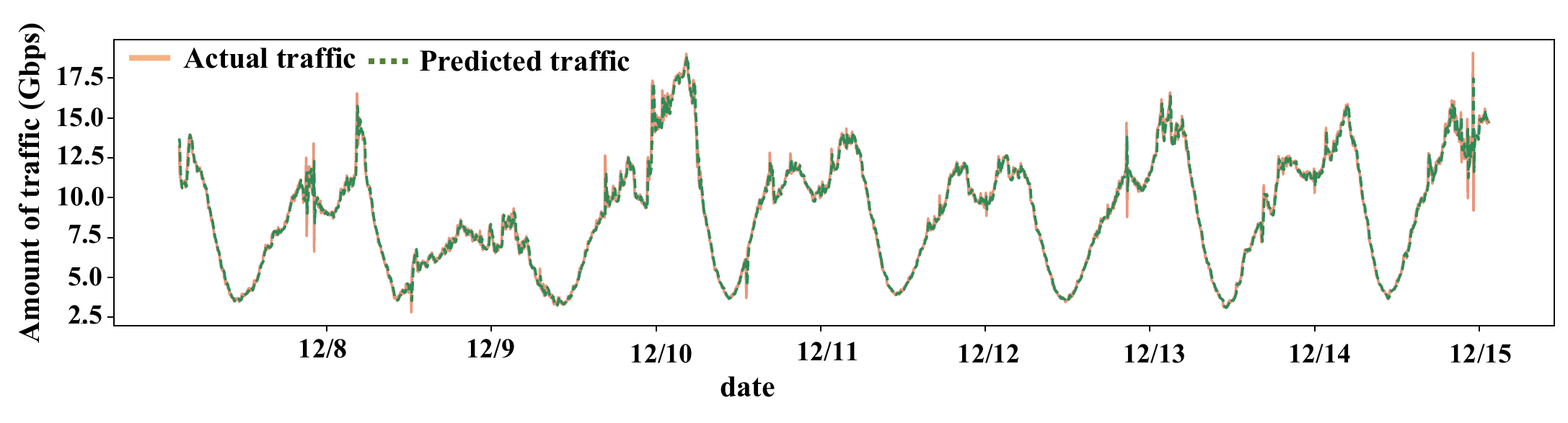}
    \caption{Actual vs. predicted traffic by LSTM\_Seq2Seq\_ATN\_EMD\_KNN.}
    \label{fig:actualvspredicted}
\end{figure*}

\subsection{Performance Analysis of Prediction Model with Noise Reduction and Outlier Mitigation}
\label{traffic denoise}
To further improve model performance, we analyze noise in our traffic data. Noise refers to random variations or errors in the data that can affect the accuracy of a model's predictions. Noise can arise from various sources, such as measurement errors, data recording issues, or other random fluctuations. By applying noise reduction techniques, such as smoothing, filtering, or denoising algorithms, the noisy data can be cleaned up, resulting in a more accurate representation of the underlying pattern in the data. This can lead to an improved model performance by reducing the impact of noise on the model's predictions. 

After removing outliers from a dataset using techniques such as KNN imputation, it is often the case that the remaining data still contain some level of noise or unwanted variability. This noise can obscure the underlying patterns and relationships in the data, making it more difficult to analyze and interpret. Therefore, it can be useful to apply additional noise reduction techniques to further improve the quality of the data. EMD-based noise reduction is one such technique we used to remove unwanted noise and variability from the data, resulting in a smoother and more interpretable signal. We summarized Noise-to-Signal (SNR) ratio in Table \ref{tab:snr} for analyzing the signal quality after denoising. The SNR values that we obtained indicate that the EMD-based noise reduction method has significantly improved the quality of the signal. A negative SNR value, -7.05 dB, for the noisy signal indicates that the noise in the signal is actually stronger than the signal itself. This can make it difficult to accurately analyze and interpret the data. However, after applying EMD-based noise reduction, the denoised signal has a much higher SNR value (21.47 dB), indicating that the quality of the signal has been significantly improved relative to the noise. This means that the denoised signal is now much easier to analyze and can provide more accurate and reliable insights into the underlying patterns and relationships in the data. In summary, the significant improvement in SNR value for the denoised signal compared to the noisy signal suggests that the EMD-based noise reduction method has been effective in reducing unwanted noise and variability in the data, resulting in a higher quality and more interpretable signal. In Fig. \ref{fig:denoise}, we depicted actual traffic, denoised traffic, and the noise in the dataset. 

Our proposed RNN\_KNN\_EMD model prediction error is lower by more than 3\% compared to RNN. This states the effectiveness of our proposed traffic prediction method in handling real-world internet traffic, which might have noise and outliers due to various external and internal factors. The experiment was expanded by including two extensions of RNN called Long Short-Term Memory (LSTM) and Gated Recurrent Unit (GRU) because these models had a greater capacity for knowledge retention from longer sequences than the RNN. Traditional LSTM and GRU outperformed RNN by 2.48\% and 1.1\% more prediction accuracy, respectively. Since RNN has an inherent problem of vanishing gradient problem in handling longer sequence data, its prediction accuracy was smaller than LSTM and GRU, specifically designed to address RNN limitations. Our experimental results indicate that LSTM and GRU have more power to retain information from sequential data compared to RNN. We further improved LSTM and GRU performance by integrating our proposed denoising and outlier detection module. Our proposed LSTM\_KNN\_EMD and GRU\_KNN\_EMD perform better than conventional LSTM and GRU. The LSTM\_KNN\_EMD prediction accuracy is increased by 1.73\% compared to LSTM, while for GRU\_KNN\_EMD, the accuracy improvement was 2.89\% than GRU. Since deep sequence models considered only the temporal information for learning, the noise and outlier in the training data ultimately affect its generalization capability resulting in lower accuracy. Our EMD-based noise reduction and empirical rule-based outlier mitigation provide traffic with random abrupt changes for model training, which eventually increase the prediction accuracy and decrease the average prediction error between actual and predicted traffic. Finally, we got our best prediction accuracy from our proposed LSTM\_Seq2Seq\_ATN\_EMD\_KNN model, where we integrated an attention layer. The extra layer helps our Seq2Seq architecture to extract strong contextual information from the traffic data. The conventional LSTM\_Seq2Seq\_ATN model without the proposed module provided the lowest prediction error of 3.95\% compared to other deep learning models. Moreover, our proposed LSTM\_Seq2Seq\_ATN\_KNN\_EMD prediction accuracy is the highest among all prediction models with 96.48\% accurate forecast. Our proposed model performs better than the traditional Seq2Seq model with nearly 1\% more prediction accuracy. Among all deep sequence model architectures, the Seq2Seq model with an extra attention layer performs better by extracting strong contextual information from longer sequence data. A comparison between actual and predicted traffic by our best-performing model has been made depicted in Fig. \ref{fig:actualvspredicted}.

In Fig. \ref{fig:outliernoise_vs_standard}, we depicted error reduction by our proposed model compared to conventional and only KNN-integrated models. We achieved better prediction accuracy for all models by combining both outlier and noise handling in the traffic data. For example, LSTM\_Seq2Seq\_KNN\_EMD gave approximately 34\% less error than LSTM\_Seq2Seq while LSTM\_Seq2Seq\_KNN provided approximately 24\% less prediction error. So, it is evident from Fig. \ref{fig:outliernoise_vs_standard} that combining outlier management and noise reduction would be a better approach to dealing with real-world internet traffic. In Fig. \ref{fig:outliernoise_vs_standard}, the model integrated with both outlier and noise performed better than models with only the outlier management module. For example, our best performing LSTM\_Seq2Seq\_KNN\_EMD model provided us approximately 11\% less error than LSTM\_Seq2Seq\_KNN. Therefore, based on all models' performance, we can conclude that traffic analysis is crucial before using traffic data to develop the model. Especially the outliers and noise in the traffic might affect the model prediction accuracy, and it is essential to deal with them first before using traffic data to train and evaluate the model.

%% file: main.bbl
\begin{thebibliography}{10}
\providecommand{\url}[1]{#1}
\csname url@samestyle\endcsname
\providecommand{\newblock}{\relax}
\providecommand{\bibinfo}[2]{#2}
\providecommand{\BIBentrySTDinterwordspacing}{\spaceskip=0pt\relax}
\providecommand{\BIBentryALTinterwordstretchfactor}{4}
\providecommand{\BIBentryALTinterwordspacing}{\spaceskip=\fontdimen2\font plus
\BIBentryALTinterwordstretchfactor\fontdimen3\font minus
  \fontdimen4\font\relax}
\providecommand{\BIBforeignlanguage}[2]{{%
\expandafter\ifx\csname l@#1\endcsname\relax
\typeout{** WARNING: IEEEtran.bst: No hyphenation pattern has been}%
\typeout{** loaded for the language `#1'. Using the pattern for}%
\typeout{** the default language instead.}%
\else
\language=\csname l@#1\endcsname
\fi
#2}}
\providecommand{\BIBdecl}{\relax}
\BIBdecl

\bibitem{intro1}
D.~Jiang, F.~Wang, Z.~Lv, S.~Mumtaz, S.~Al-Rubaye, A.~Tsourdos, and O.~Dobre,
  ``Qoe-aware efficient content distribution scheme for satellite-terrestrial
  networks,'' \emph{IEEE Transactions on Mobile Computing}, 2021.

\bibitem{intro2}
D.~Jiang, Z.~Wang, L.~Huo, and S.~Xie, ``A performance measurement and analysis
  method for software-defined networking of iov,'' \emph{IEEE Transactions on
  Intelligent Transportation Systems}, vol.~22, no.~6, pp. 3707--3719, 2020.

\bibitem{intro4}
D.~Jiang, W.~Wang, L.~Shi, and H.~Song, ``A compressive sensing-based approach
  to end-to-end network traffic reconstruction,'' \emph{IEEE Transactions on
  Network Science and Engineering}, vol.~7, no.~1, pp. 507--519, 2018.

\bibitem{ciscoCiscoAnnual}
``{C}isco {A}nnual {I}nternet {R}eport - {C}isco {A}nnual {I}nternet {R}eport
  (2018–2023) {W}hite {P}aper --- cisco.com,''
  \url{https://www.cisco.com/c/en/us/solutions/collateral/executive-perspectives/annual-internet-report/white-paper-c11-741490.html},
  [Accessed 15-Mar-2023].

\bibitem{alghamdi2019forecasting}
T.~Alghamdi, K.~Elgazzar, M.~Bayoumi, T.~Sharaf, and S.~Shah, ``Forecasting
  traffic congestion using arima modeling,'' in \emph{2019 15th international
  wireless communications \& mobile computing conference (IWCMC)}.\hskip 1em
  plus 0.5em minus 0.4em\relax IEEE, 2019, pp. 1227--1232.

\bibitem{tang2019traffic}
J.~Tang, X.~Chen, Z.~Hu, F.~Zong, C.~Han, and L.~Li, ``Traffic flow prediction
  based on combination of support vector machine and data denoising schemes,''
  \emph{Physica A: Statistical Mechanics and its Applications}, vol. 534, p.
  120642, 2019.

\bibitem{jaffry2020cellular}
S.~Jaffry and S.~F. Hasan, ``Cellular traffic prediction using recurrent neural
  networks,'' in \emph{2020 IEEE 5th International Symposium on
  Telecommunication Technologies (ISTT)}.\hskip 1em plus 0.5em minus
  0.4em\relax IEEE, 2020, pp. 94--98.

\bibitem{intro6}
D.~Jiang, Y.~Wang, Z.~Lv, S.~Qi, and S.~Singh, ``Big data analysis based
  network behavior insight of cellular networks for industry 4.0
  applications,'' \emph{IEEE Transactions on Industrial Informatics}, vol.~16,
  no.~2, pp. 1310--1320, 2019.

\bibitem{liter1}
X.~Fan, Y.~Wang, and M.~Zhang, ``Network traffic forecasting model based on
  long-term intuitionistic fuzzy time series,'' \emph{Information sciences},
  vol. 506, pp. 131--147, 2020.

\bibitem{yang2021network}
H.~Yang, X.~Li, W.~Qiang, Y.~Zhao, W.~Zhang, and C.~Tang, ``A network traffic
  forecasting method based on sa optimized arima--bp neural network,''
  \emph{Computer Networks}, vol. 193, p. 108102, 2021.

\bibitem{yang2021farima}
J.~Yang, H.~Sheng, H.~Wan, and F.~Yu, ``Farima model based on particle
  swarm-genetic hybrid algorithm optimization and application,'' in \emph{2021
  3rd International Academic Exchange Conference on Science and Technology
  Innovation (IAECST)}.\hskip 1em plus 0.5em minus 0.4em\relax IEEE, 2021, pp.
  188--192.

\bibitem{sarima}
P.~Kromkowski, S.~Li, W.~Zhao, B.~Abraham, A.~Osborne, and D.~E. Brown,
  ``Evaluating statistical models for network traffic anomaly detection,'' in
  \emph{2019 Systems and Information Engineering Design Symposium (SIEDS)},
  2019, pp. 1--6.

\bibitem{liter11}
Y.~Wu, Y.~Cui, W.~Yu, C.~Lu, and W.~Zhao, ``Modeling and forecasting of
  timescale network traffic dynamics in m2m communications,'' in \emph{2019
  IEEE 39th International Conference on Distributed Computing Systems
  (ICDCS)}.\hskip 1em plus 0.5em minus 0.4em\relax IEEE, 2019, pp. 711--721.

\bibitem{liter2}
C.~Katris and S.~Daskalaki, ``Dynamic bandwidth allocation for video traffic
  using farima-based forecasting models,'' \emph{Journal of Network and Systems
  Management}, vol.~27, no.~1, pp. 39--65, 2019.

\bibitem{sheng2020alpha}
H.~Sheng, Q.~Yan, and K.~Li, ``Alpha stable distribution based farima modeling
  and forecasting for network traffic data,'' in \emph{Journal of Physics:
  Conference Series}, vol. 1574, no.~1.\hskip 1em plus 0.5em minus 0.4em\relax
  IOP Publishing, 2020, p. 012135.

\bibitem{christian2021network}
G.~A. Christian, I.~P. Wijaya, and R.~F. Sari, ``Network traffic prediction of
  mobile backhaul capacity using time series forecasting,'' in \emph{2021
  International Seminar on Intelligent Technology and Its Applications
  (ISITIA)}.\hskip 1em plus 0.5em minus 0.4em\relax IEEE, 2021, pp. 58--62.

\bibitem{liter4}
J.~Zhou, X.~Yang, L.~Sun, C.~Han, and F.~Xiao, ``Network traffic prediction
  method based on improved echo state network,'' \emph{IEEE Access}, vol.~6,
  pp. 70\,625--70\,632, 2018.

\bibitem{hou2018fuzzy}
Y.~Hou, L.~Zhao, and H.~Lu, ``Fuzzy neural network optimization and network
  traffic forecasting based on improved differential evolution,'' \emph{Future
  Generation Computer Systems}, vol.~81, pp. 425--432, 2018.

\bibitem{liter5}
W.~Zhang and D.~Wei, ``Prediction for network traffic of radial basis function
  neural network model based on improved particle swarm optimization
  algorithm,'' \emph{Neural Computing and Applications}, vol.~29, no.~4, pp.
  1143--1152, 2018.

\bibitem{8936255}
O.~A. Adeleke, ``Echo-state networks for network traffic prediction,'' in
  \emph{2019 IEEE 10th Annual Information Technology, Electronics and Mobile
  Communication Conference (IEMCON)}, 2019, pp. 0202--0206.

\bibitem{zhou2022time}
Y.~Zhou, M.~Zhang, and K.-P. Lin, ``Time series forecasting by the novel
  gaussian process wavelet self-join adjacent-feedback loop reservoir model,''
  \emph{Expert Systems with Applications}, vol. 198, p. 116772, 2022.

\bibitem{li2019network}
H.~Li, ``Network traffic prediction of the optimized bp neural network based on
  glowworm swarm algorithm,'' \emph{Systems Science \& Control Engineering},
  vol.~7, no.~2, pp. 64--70, 2019.

\bibitem{li2006algorithm}
W.~Li and Y.~Hori, ``An algorithm for extracting fuzzy rules based on rbf
  neural network,'' \emph{IEEE Transactions on Industrial Electronics},
  vol.~53, no.~4, pp. 1269--1276, 2006.

\bibitem{liter3}
Q.~Yang, W.~Hao, L.~Ge, W.~Ruan, and F.~Chi, ``Farima model-based communication
  traffic anomaly detection in intelligent electric power substations,''
  \emph{IET Cyber-Physical Systems: Theory \& Applications}, vol.~4, no.~1, pp.
  22--29, 2019.

\bibitem{huang2018backbone}
W.~Huang, J.~Zhang, S.~Liang, and H.~Sun, ``Backbone network traffic prediction
  based on modified eemd and quantum neural network,'' \emph{Wireless Personal
  Communications}, vol.~99, pp. 1569--1588, 2018.

\bibitem{littable1}
P.~Kromkowski, S.~Li, W.~Zhao, B.~Abraham, A.~Osborne, and D.~E. Brown,
  ``Evaluating statistical models for network traffic anomaly detection,'' in
  \emph{2019 Systems and Information Engineering Design Symposium (SIEDS)},
  2019, pp. 1--6.

\bibitem{liter8}
S.~Zhang, L.~Zhou, X.~Chen, L.~Zhang, L.~Li, and M.~Li, ``Network-wide traffic
  speed forecasting: 3d convolutional neural network with ensemble empirical
  mode decomposition,'' \emph{Computer-Aided Civil and Infrastructure
  Engineering}, vol.~35, no.~10, pp. 1132--1147, 2020.

\bibitem{narejo2018application}
S.~Narejo and E.~Pasero, ``An application of internet traffic prediction with
  deep neural network,'' \emph{Multidisciplinary Approaches to Neural
  Computing}, pp. 139--149, 2018.

\bibitem{liter7}
X.~Han and F.~Qi, ``Network traffic forecasting using ifa-lstm,'' in
  \emph{International Conference on Computer Engineering and Networks}.\hskip
  1em plus 0.5em minus 0.4em\relax Springer, 2018, pp. 681--692.

\bibitem{wang2018network}
W.~Wang, Y.~Bai, C.~Yu, Y.~Gu, P.~Feng, X.~Wang, and R.~Wang, ``A network
  traffic flow prediction with deep learning approach for large-scale
  metropolitan area network,'' in \emph{NOMS 2018-2018 IEEE/IFIP Network
  Operations and Management Symposium}.\hskip 1em plus 0.5em minus 0.4em\relax
  IEEE, 2018, pp. 1--9.

\bibitem{liter10}
B.~A. Pratomo, P.~Burnap, and G.~Theodorakopoulos, ``Blatta: Early exploit
  detection on network traffic with recurrent neural networks,'' \emph{Security
  and Communication Networks}, vol. 2020, 2020.

\bibitem{9146846}
Y.~Liu, S.~Garg, J.~Nie, Y.~Zhang, Z.~Xiong, J.~Kang, and M.~S. Hossain, ``Deep
  anomaly detection for time-series data in industrial iot: A
  communication-efficient on-device federated learning approach,'' \emph{IEEE
  Internet of Things Journal}, vol.~8, no.~8, pp. 6348--6358, 2021.

\bibitem{li2020traffic}
Y.~Li and W.~Tu, ``Traffic modelling for iot networks: A survey,'' in
  \emph{Proceedings of the 10th International Conference on Information
  Communication and Management}, 2020, pp. 4--9.

\bibitem{improved-sae}
P.~Li, Z.~Chen, L.~T. Yang, J.~Gao, Q.~Zhang, and M.~J. Deen, ``An improved
  stacked auto-encoder for network traffic flow classification,'' \emph{IEEE
  Network}, vol.~32, no.~6, pp. 22--27, 2018.

\bibitem{wei2017network}
D.~Wei, ``Network traffic prediction based on rbf neural network optimized by
  improved gravitation search algorithm,'' \emph{Neural Computing and
  Applications}, vol.~28, pp. 2303--2312, 2017.

\bibitem{liter6}
K.~Zhang, Z.~Hu, X.-T. Gan, and J.-B. Fang, ``A network traffic prediction
  model based on quantum-behaved particle swarm optimization algorithm and
  fuzzy wavelet neural network,'' \emph{Discrete Dynamics in Nature and
  Society}, vol. 2016, 2016.

\bibitem{zhou2022self}
H.~Zhou, Y.~Li, H.~Xu, Y.~Su, and L.~Chen, ``A self-organizing fuzzy neural
  network modeling approach using an adaptive quantum particle swarm
  optimization,'' \emph{Applied Intelligence}, pp. 1--24, 2022.

\bibitem{method1}
B.~Premanode, J.~Vongprasert, and C.~Toumazou, ``Prediction of exchange rates
  using averaging intrinsic mode function and multiclass support vector
  regression.'' \emph{Artif. Intell. Res.}, vol.~2, no.~2, pp. 47--61, 2013.

\bibitem{method2}
M.~R. Chernick, \emph{The essentials of biostatistics for physicians, nurses,
  and clinicians}.\hskip 1em plus 0.5em minus 0.4em\relax John Wiley \& Sons,
  2011.

\bibitem{chollet2015keras}
F.~Chollet \emph{et~al.}, ``Keras,'' \url{https://keras.io}, 2015.

\end{thebibliography}
